\documentclass{article}
\usepackage{glossaries}
\usepackage{graphicx}
\usepackage[ruled,vlined]{algorithm2e}
\usepackage{array}
\usepackage{comment}
\usepackage[center]{caption}
\usepackage{subcaption}
\usepackage{color}

\newcolumntype{C}[1]{>{\centering\let\newline\\\arraybackslash\hspace{0pt}}m{#1}}
\newcommand\T{\rule{0pt}{2.6ex}}       
\newcommand\B{\rule[-1.8ex]{0pt}{0pt}} 

\definecolor{mygray}{RGB}{200, 200, 200}
\definecolor{mygreen}{RGB}{1, 200, 1}
\definecolor{lightgray}{gray}{0.8}
\definecolor{lightlightgray}{gray}{0.9}

\usepackage[colorinlistoftodos,prependcaption,textsize=tiny]{todonotes}

\usepackage[final]{corl_2020} 

\title{Interactive Imitation Learning in State-Space}
\author{
	Snehal Jauhri \qquad{} Carlos Celemin \qquad{} Jens Kober\\
	Department of Cognitive Robotics\\
	Delft University of Technology, Netherlands\\
	\texttt{snehal.jauhri@gmail.com \quad{} \texttt{\{c.e.celeminpaez, j.kober\}@tudelft.nl}}
}

\begin{document}
	\maketitle
	\vspace{-15pt}
	\begin{abstract}
		Imitation Learning techniques enable programming the behavior of agents through demonstrations rather than manual engineering. However, they are limited by the quality of available demonstration data. Interactive Imitation Learning techniques can improve the efficacy of learning since they involve teachers providing feedback while the agent executes its task. In this work, we propose a novel Interactive Learning technique that uses human feedback in state-space to train and improve agent behavior (as opposed to alternative methods that use feedback in action-space). Our method titled Teaching Imitative Policies in State-space~(TIPS) enables providing guidance to the agent in terms of `changing its state' which is often more intuitive for a human demonstrator. 
		Through continuous improvement via corrective feedback, agents trained by non-expert demonstrators using TIPS outperformed the demonstrator and conventional Imitation Learning agents. 
	\end{abstract}
	
	
	\vspace{-7pt}
	\section{Introduction}
	\vspace{-3pt}
	\label{sec:intro}
	Imitation Learning (IL) is a machine learning technique in which an agent learns to perform a task using example demonstrations \citep{Osa2018a}. 
	This eliminates the need for humans to pre-program the required behavior  for a task, instead utilizing the more intuitive mechanism of demonstrating it \citep{Osa2018a}. 
	Advancements in Imitation Learning techniques have led to successes in learning tasks such as robot locomotion \citep{Zucker2011}, helicopter flight \citep{Abbeel2010} and learning to play games \citep{Silver2016}. 
	There have also been research efforts to make training easier for demonstrators. 
	This is done by allowing them to interact with the agent by providing feedback as it performs the task, also known as Interactive IL  \citep{Argall2008, Chernova2009, Celemin2019a}.
	
	One limitation of current IL and Interactive IL techniques is that they typically require demonstrations or feedback in the \emph{action-space} of the agent. Humans commonly learn behaviors by understanding the required \emph{state} transitions of a task, not the precise actions to be taken \citep{Liu2018}. 
	Additionally, providing demonstration or feedback in the action-space can be difficult for demonstrators. For instance, teaching a robotic arm manipulation task with joint level actions (motor commands) requires considerable demonstrator expertise. It would be easier to instead provide \emph{state-space} information such as the Cartesian position of the end effector or the object to be manipulated (e.g., moving towards/away from the object). Considering cases where a tool is attached to the robot arm, feedback could also be provided on how the tool interacts with the environment (e.g., tightening/loosening the grasp of an object).
	
	
	In this paper, a novel Interactive Learning method is proposed that utilizes feedback in state-space to learn behaviors. 
	The performance of the proposed method (TIPS) is evaluated for various control tasks as part of the OpenAI Gym toolkit and for manipulation tasks using a KUKA LBR iiwa robot arm. 
	Although it requires an additional dynamics learning step, the method compares favorably to other Imitation and Interactive Learning methods in non-expert demonstration scenarios.
	
	\section{Related Work}\label{sec:background}
	
	In recent literature, several Interactive Imitation Learning methods have been proposed that enable demonstrators to guide agents by providing corrective action labels \citep{Ross2011}, corrective feedback \citep{Argall2011a, Celemin2019a} or evaluative feedback \citep{Knox2009, Christiano2017}. 
	For non-expert demonstrators, providing corrective feedback in the form of adjustments to the current states/actions being visited/executed by the agent is easier than providing exact state/action labels \cite{Celemin2019a}. Moreover, evaluative feedback methods require demonstrators to score good and bad behaviors, which could be ambiguous when scoring multiple sub-optimal agent behaviors.
	
	Among corrective feedback learning techniques, a typical approach is to utilize corrections in the action-space \citep{Celemin2019a,Perez2019} or to use predefined advice-operators \citep{Argall2008, Argall2011a} to guide agents. 
	However, providing feedback in the action-space is often not intuitive for demonstrators (e.g., action-space as joint torques or angles of a robotic arm). 
	Further, defining advice-operators requires significant prior knowledge about the environment as well as the task to be performed, thus limiting the generalizability of such methods. This work proposes an alternative approach of using corrective feedback in state-space to advise agents.
	
	There has been recent interest in Imitation Learning methods that learn using state/observation information only.
	This problem is termed as Imitation from Observation (IfO) and enables learning from state trajectories of humans performing the task. 
	To compute the requisite actions, many IfO methods propose using a learnt Inverse Dynamics Model (IDM) \citep{Nair2017,Torabi2018} which maps state transitions to the actions that produce those state transitions. 
	However, teaching agents using human interaction in an IfO setting has not been studied. 
	
	In our approach, we combine the concept of state transition to action mapping by learning inverse dynamics with an Interactive Learning framework. The demonstrator provides state-space corrective feedback to guide the agent's behavior towards desired states.
	Meanwhile, an inverse dynamics scheme is used to ensure the availability of the requisite actions to learn the policy.

	
	\section{Teaching Imitative Policies in State-space (TIPS)}
	\label{sec:method}
	
	The principle of TIPS is to allow the agent to execute its policy while a human demonstrator observes and suggests modifications to the state visited by the agent at any given time. 
	This feedback is advised and used to update the agent’s policy online, i.e., during the execution itself.
	
	\subsection{Corrective Feedback}\label{sub::errorassumption}
	Human feedback ($h_t$, at time step $t$) is in the form of binary signals implying an increase/decrease in the value of a state (i.e., $h_t \in \{-1, 0, +1\}$, where zero implies no feedback). 
	Each dimension of the state has a corresponding feedback signal. 
	The assumption is that non-expert human demonstrators, who may not be able to provide accurate correction values could still provide binary signals which show the trend of state modification. 
	To convert these signals to a modification value, an error constant hyper-parameter $e$ is chosen for each state dimension. 
	Thus, the human desired state ($s^{des}_{t+1}$) is computed as:
	\begin{equation}
		s^{des}_{t+1} = s_t + h_t \cdot e.
	\end{equation}
	
	The feedback ($h_t$), error constant $e$, and the desired modification can be both in the full state or partial state. Thus, the demonstrator is allowed to only suggest modifications in the partial state dimensions that are well understood or easy to observe for the demonstrator. 
	Moreover, even though the change in state computed using binary feedback may be larger/smaller than what the demonstrator is suggesting, previous methods \cite{Celemin2019a,Perez2019} have shown that it is sufficient to capture the \emph{trend} of modification. 
	If a sequence of feedback provided in a state is in the same direction (increase/decrease), the demonstrator is suggesting a large magnitude change. 
	Conversely, if the feedback alternates between increase/decrease, a smaller change around a set-point is suggested \citep{Celemin2019a}.
	To obtain this effect when updating the policy, information from past feedback is also used via a replay memory mechanism as in \cite{Perez2019,Perez2020}.
	
	\begin{figure}[t!]
		\centering
		\includegraphics[scale=0.5]{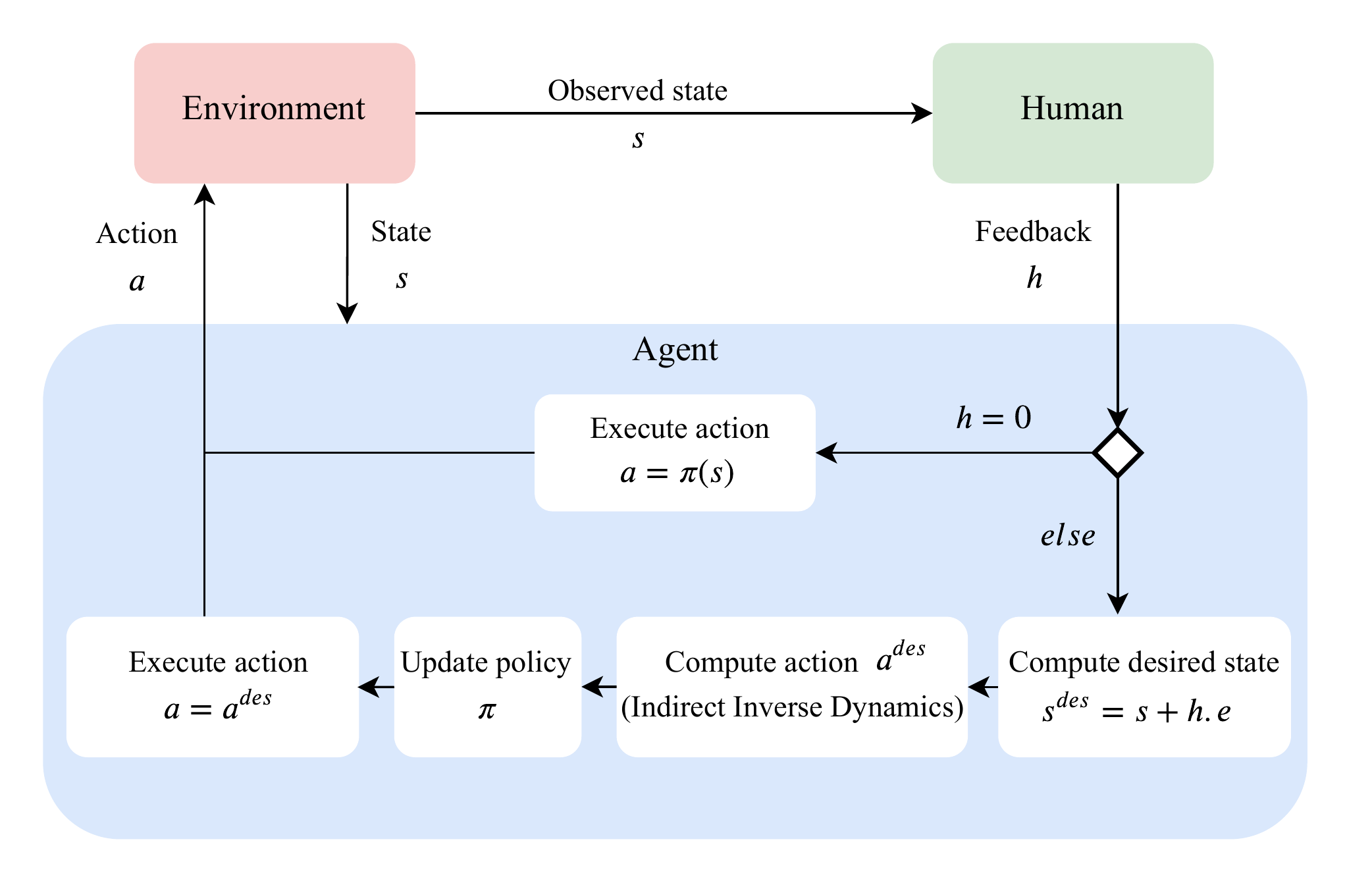}
		\captionsetup{justification=centering, font=footnotesize}
		\caption{High-level representation of the learning framework of TIPS}
		\label{fig::TIPS}
	\end{figure}
	
	\subsection{Mapping state transitions to actions}\label{sub::IDM}
	To realize the transition from the current state to the desired state, i.e., $s_t \to s^{des}_{t+1}$, an appropriate action ($a^{des}_t$) needs to be computed.
	For this, some methods have proposed using or learning an Inverse Dynamics Model (IDM) \citep{Nair2017,Torabi2018}. 
	In this work we assume that an IDM is not already available, which can be the case in environments with dynamics that are unknown or difficult to model.
	Moreover, IDMs are ill-suited in our case for two main reasons. 
	Firstly, the feedback provided by the demonstrator can be in the partial state-dimension, leading to ambiguity regarding the desired state transition in the remaining dimensions. 
	Secondly, the desired state transition ($s_t \to s^{des}_{t+1}$) may be infeasible. 
	There may not exist an action that leads to the human suggested state transition in a single time step.
	
	We propose to instead use an indirect inverse dynamics method to compute requisite actions.
	Possible actions are sampled ($a \in A$) and a learnt Forward Dynamics Model (FDM) ($f$) is used to predict the next states ($\hat{s}_{t+1} = f(s_t,a)$) for these actions.
	The action that results in a subsequent state that is closest to the desired state is chosen. The desired and predicted states can be in the full or partial state dimensions. Mathematically, we can write the action computation as: 
	\begin{equation}
		a^{des}_{t} = \underset{a}{\text{arg min}} \left\Vert f(s_t,a) - s^{des}_{t+1} \right\Vert,
	\end{equation}
	where $a \in A$ with $N_a$ uniform samples.
	
	\subsection{Training Mechanism}\label{sub::algorithm}
	We represent the policy $\pi(s)$ using a feed-forward artificial neural network and use a training mechanism inspired by D-COACH \citep{Perez2019}.
	This involves an immediate training step using the current state-action sample as well as a training step using a batch sampled from a demonstration replay memory.
	Lastly, to ensure sufficient learning iterations to train the neural network, a batch replay training step is also carried out periodically every $T_{update}$ time-steps.
	
	Crucially, the computed action $a^{des}_t$ is also executed immediately by the agent.
	This helps speed up the learning process since further feedback can be received in the demonstrator requested state to learn the next action to be taken. 
	The overall learning framework of TIPS can be seen in Figure~\ref{fig::TIPS}. 
	
	The overall TIPS method consists of two phases:
	\begin{itemize}
		\item In an initial model-learning phase, samples are generated by executing an exploration policy $\pi_e$ (random policy implementation) and used to learn an initial FDM $f_{\theta}$. The samples are added to an experience buffer $E$ that is used later when updating the model.
		
		\item In the teaching phase, the policy $\pi_{\phi}$ is trained using an immediate update step every time feedback is advised as well as a periodic update step using past feedback from a demonstration buffer $D$. Moreover, to improve the FDM, it is trained after every episode using the consolidated new and previous experience gathered in $E$.
	\end{itemize}
	
	The pseudo-code of TIPS can be seen in Algorithm~\ref{algo::main}. 
	In our implementation of TIPS (\href{https://github.com/sjauhri/Interactive-Learning-in-State-space}{github.com/sjauhri/Interactive-Learning-in-State-space}), the FDM and policy are represented using neural networks and the Adam variant of stochastic gradient descent~\citep{Kingma2014} is used for training. 
	
	\begin{algorithm}[h]
		\SetAlgoLined
		\vspace{4pt}
		\Indm
		\textbf{\underline{Initial Model-Learning Phase:}}\\
		\vspace{4pt}
		\Indp
		Generate $N_e$ experience samples $\{s_i,a_i\}^{N_e}_{1}$ by executing a random/exploration policy $\pi_e$\\
		Append samples to experience buffer $E$\\
		Learn forward dynamics model $f_{\theta}$ using inputs $\{s_i,a_i\}^{N_e}_{1}$ and targets $\{s_{i+1}\}^{N_e}_{1}$\\
		\Indm
		\hrulefill\\
		\textbf{\underline{Teaching Phase:}}\\
		\vspace{4pt}
		\Indp
		\For{episodes}{
			\For{$t = 0,1,2,\ldots,T$}{
				Visit state $s_t$\\
				Get human corrective feedback $h_t$\\
				\eIf{$h_t \operatorname{is} \operatorname{not} \operatorname{} 0$}{
					Compute desired state $s^{des}_{t+1} = s_t +$ $h_t \cdot e$\\
					Compute action $a^{des}_{t} = \underset{a}{\text{arg min}} \left\Vert f_{\theta}(s_t,a) - s^{des}_{t+1} \right\Vert$, using $N_a$ sampled actions\\
					Append $(s_t, a^{des}_{t})$ to demonstration buffer $D$\\
					Update policy $\pi_{\phi}$ using pair $(s_t, a^{des}_{t})$ and using batch sampled from $D$\\
					Execute action $a_t = a^{des}_{t}$, reach state $s_{t+1}$\\
				}
				{
					\textit{\#\# No feedback}\\
					Execute action $a_t = \pi_{\phi}(s_t)$, reach state $s_{t+1}$\\
				}
				Append $(s_t, a_t, s_{t+1})$ to experience buffer $E$\\
				\If{$\operatorname{mod}(t, T_{update})$}{
					Update policy $\pi_{\phi}$ using batch sampled from demonstration buffer $D$\\
				}
				
			}
			Update learnt FDM $f_{\theta}$ using samples from experience buffer $E$\\
		}
		\Indm
		\caption{Teaching Imitative Policies in State-space (TIPS)}
		\label{algo::main}
	\end{algorithm}
	
	\section{Experimental Setting}
	\label{sec:experimental}
	Experiments are set up to evaluate TIPS and compare it to other methods when teaching simulated tasks with non-expert human participants as demonstrators (Section~\ref{sec:eval}). 
	We also validate the method on a real robot by designing two manipulation tasks with a robotic arm (Section~\ref{sec:validate}).
	
	
	\subsection{Evaluation}
	\label{sec:eval}
	For the evaluation of TIPS, we use three simulated tasks from the OpenAI gym toolkit \citep{Brockman2016}, namely: CartPole, Reacher and LunarLanderContinuous. A simplified version of the Reacher task with a fixed target position is used. 
	The cumulative reward obtained by the agent during execution is used as a performance metric. 
	The parameter settings for each of the domains/tasks in the experiments can be seen in Table~\ref{table:hyper}. Notably, given the small dimensionality of the action spaces in our settings, the evaluation of action samples ($N_a$) is computationally inexpensive and almost instantaneous.
	
	The performance of a TIPS agent is compared against the demonstrator's own performance when executing the task via tele-operation, and against other agents trained via IL techniques using the tele-operation data. 
	It is also of interest to highlight the differences between demonstration in state-space versus action-space. 
	For this, both tele-operation and corrective feedback learning techniques in state and action spaces are compared. The comparison is with IL methods and not IfO methods (such as~\citep{Torabi2018}) since IfO methods assume no knowledge of actions during tele-operation. This is not true in our interactive learning setting where actions are known but only the interface can differ.
	
	The following techniques are used for comparison. 
	\begin{itemize}
		\item \textbf{Tele-operation in Action-space:} Demonstrator executes task using action commands. 
		
		\item \textbf{Tele-operation in State-space:} Demonstrator executes task by providing state-space information (as per Table~\ref{table:hyper}) with actions computed using inverse dynamics in a similar way as TIPS.
		
		\item \textbf{Behavioral Cloning (BC):} Supervised learning to imitate the demonstrator using state-action demonstration data recorded during tele-operation. (Only successful demonstrations are used, i.e., those with a return of at least 40\% in the min-max range).
		
		\item \textbf{Generative Adversarial Imitation Learning (GAIL) \citep{Ho2016}:} Method that uses adversarial learning to learn a reward function and policy. Similar to BC, the successful state-action demonstration data is used for imitation. GAIL implementation by \citet{stable-baselines} is used.
		
		\item \textbf{D-COACH \citep{Perez2019}:} Interactive IL method that uses binary corrective feedback in the action space. The demonstrator suggests modifications to the current actions being executed to train the agent as it executes the task.
	\end{itemize}
	
	\begin{table}[t!]
		\captionsetup{justification=centering, font=footnotesize}
		\caption{Parameter settings in the implementation of TIPS for different tasks}
		\vspace{4pt}
		\centering
		\resizebox{\textwidth}{!}{%
			\begin{tabular}{|l  C{2.35cm}  C{2.35cm}  C{2.35cm} C {2.35cm} C{2.35cm}|}
				\hline
				\rule{0pt}{3.6ex}
				& CartPole & Reacher & LunarLander & Robot-Fishing & Robot-Laser Drawing\\
				\hline
				\T Number of exploration samples ($N_e$)& 500 & 10000 & 20000 & 4000 & 4000\\ \B
				States for feedback & Pole tip & x-y position & Vertical, angular & x-z position & x-y position\\ [-1ex]
				& position & of end effector & position & of end effector & of laser point\\
				\T Action-space dimensions & 2 (Discrete) & 2 (Continuous) & 2 (Continuous) & 2 (Continuous) & 2 (Continuous)\\
				\T Error constant ($e$) & 0.1 & 0.008 & 0.15 & 0.05 & 0.02\\
				\T Number of action samples ($N_a$) & 10 & 500 & 500 & 1000 & 1000\\
				\T Periodic policy update interval ($T_{update}$)& 10 & 10 & 10 & 10 & 10\\
				
				\T FDM Network ($f_{\theta}$) layer sizes  & 16, 16 & 64, 64 & 64, 64 & 32, 32 & 32, 32\\
				\T Policy Network ($\pi_{\phi}$) layer sizes  & 16, 16 & 32, 32 & 32, 32 & 32, 32 & 32, 32\\		
				\T Learning rate & 0.005 & 0.005 & 0.005  & 0.005 & 0.005\\ 
				\T Batch size  & 16 & 32 & 32 & 32 & 32 \\
				\hline		
		\end{tabular} }
		
		\label{table:hyper}
	\end{table}
	
	Experiments were run with non-expert human participants (age group 25-30 years) who have no prior knowledge of the tasks. A total of 22 sets of trials are performed (8, 8 and 6 participants for the CartPole, Reacher, and LunarLander tasks respectively). Participants performed four experiments: Tele-operation in action-space and state-space, training an agent using D-COACH and training an agent using TIPS. 
	To compensate for learning effect, the order of the experiments {was} changed for every participant. 
	Participants used a keyboard input interface to provide demonstration/feedback to the agent. 
	When performing tele-operation, the demonstrated actions and the corresponding states were recorded. 
	Tele-operation was deemed to be complete once no new demonstrative information could be provided (an average of 20 episodes for CartPole and Reacher, and 25 episodes for LunarLander).
	When training interactively using D-COACH and TIPS, the demonstrators provided feedback until no more agent performance improvement was observed. 
	
	To compare the demonstrator's task load, participants were also asked to fill out the NASA Task Load Index Questionnaire \citep{TLX} after each experiment.
	\vspace{-3pt}
	\subsection{Validation tasks on robot}
	\label{sec:validate}
	For the validation of TIPS on a real robot, two manipulation tasks were designed: `Fishing' and `Laser Drawing'. The tasks were performed with a velocity controlled KUKA LBR iiwa 7 robot.
	
	In the Fishing task (Figure~\ref{fig:kukaseq}), a ball is attached to the end-effector of the robot by a thread, and the objective is to move a swinging ball into a nearby cup (similar to placing a bait attached to a fishing rod). 
	To reduce the complexity of the task, the movement of the robot end-effector (and ball) is restricted to a 2-D x-z Cartesian plane. 
	To teach the task using TIPS, a keyboard interface is used to provide feedback in the x-z Cartesian robot end-effector position. 
	A learnt forward dynamics model is used to predict the position of the end-effector based on the joint commands (actions) requested to the robot. 
	To measure task performance, a reward function is defined which penalizes large actions as well as the distance ($\operatorname{\textit{dist}}$) between the ball and the center of the cup ($r_t = -\|a_t\|-\|\operatorname{\textit{dist}}_t\|$).

	In the Laser Drawing task (Figure~\ref{fig:kukalaser}), a laser pointer attached to the robot's end-effector is used to `draw' characters on a whiteboard (i.e. move the camera-tracked laser point in a desired trajectory) by moving two of the robot's joints (3rd and 5th). To teach the task, feedback is provided in the x-y position of the laser point {in the plane of the whiteboard}. In this case, learning the dynamics/kinematics of just the robot joints is insufficient. We thus  learn a forward dynamics model that predicts the position of the laser point on the whiteboard based on the joint commands (actions), but with coordinates in the frame of the whiteboard image observed by the camera.
	The reward function used to measure task performance is based on the Hausdorff distance \citep{Huttenlocher1993} between the shape drawn by the robot and a reference shape/trajectory.
	
	Note that since these experiments are run only to validate the application of TIPS to a real system, comparisons are not made with other learning methods.
	
	\vspace{-5pt}
	\section{Results}
	\label{sec:result}
	
	\begin{figure}[t]
		\captionsetup{justification=centering}	
		\begin{subfigure}[b]{0.45\textwidth}
			\centering
			\includegraphics[scale=0.4]{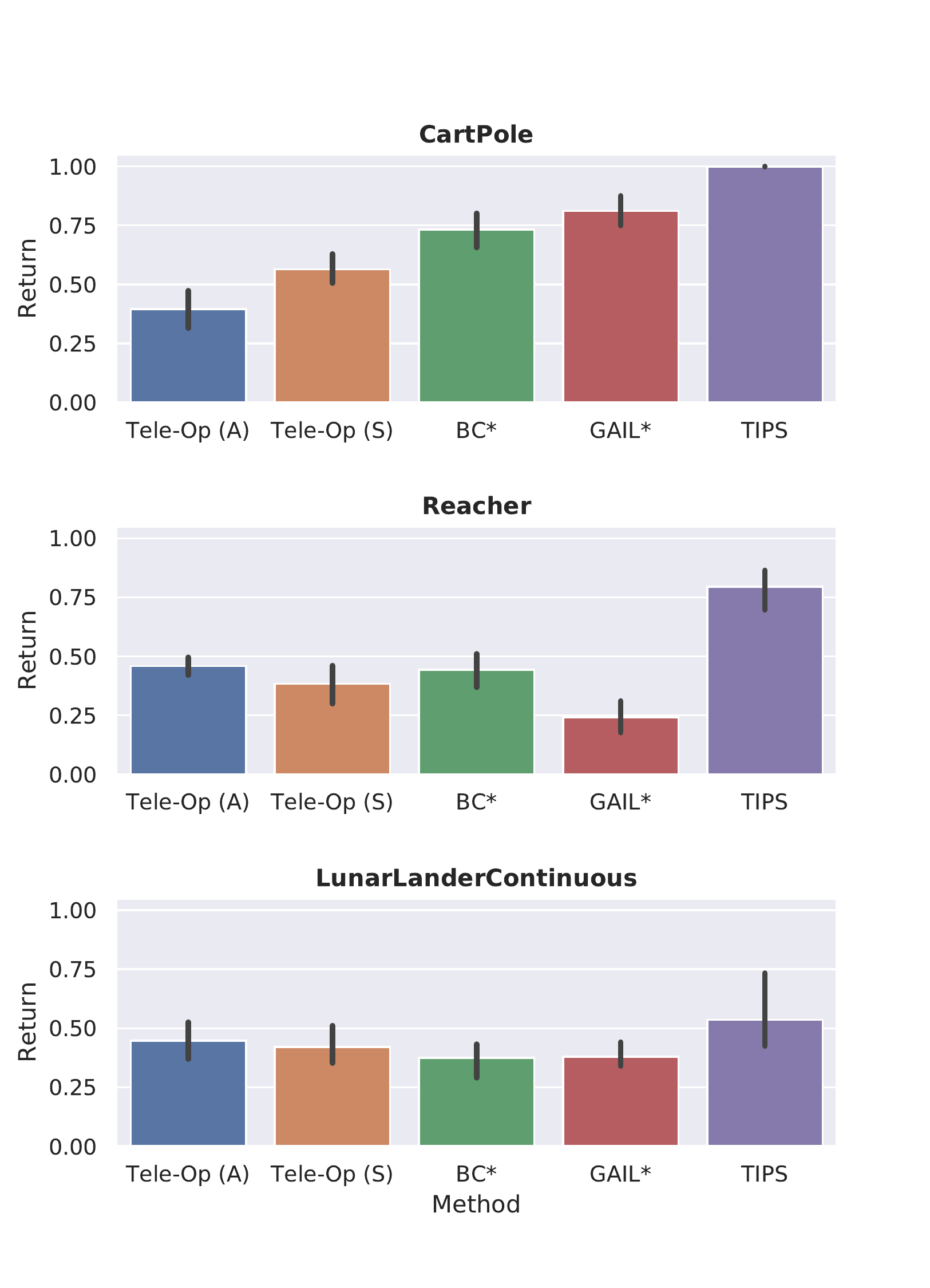}
			\captionsetup{justification=centering, font=footnotesize}
			\caption{}
			\label{fig:il_agents}
		\end{subfigure}	
		\begin{subfigure}[b]{0.45\textwidth}
			\includegraphics[scale=0.4]{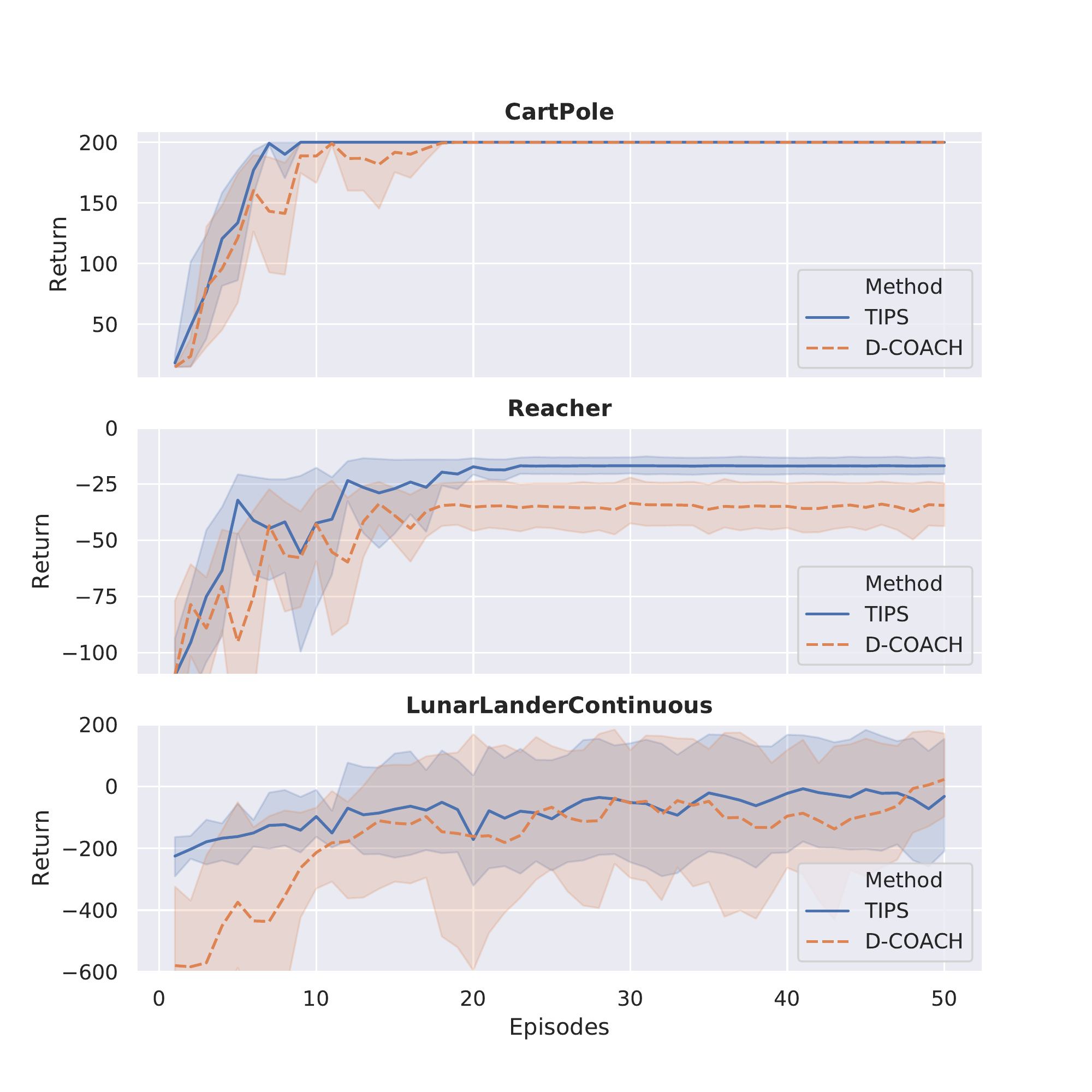}
			\hfill
			\captionsetup{justification=centering, font=footnotesize}
			\caption{}
			\label{fig:tipsvdcoach}
		\end{subfigure}
		\captionsetup{justification=justified, font=footnotesize}
		\caption{Evaluation results of TIPS. (a) Performance comparison of Tele-Operation (Action space), Tele-Operation (State space), BC, GAIL and TIPS. The return is normalized over the maximum possible return for each environment and averaged over multiple episodes and over all participants. BC* and GAIL* use only successful tele-operation data (return of at least 40\% in the normalized range). (b)	Performance of TIPS (state-space feedback) and D-COACH (action-space feedback) agents over training episodes.\vspace{-10pt}}
		\label{fig:cons}
	\end{figure}
	
	\subsection{Evaluation}
	\textbf{Performance:} Figure~\ref{fig:il_agents} shows the performance obtained for the tasks (averaged over all participants) using tele-operation, agents trained via {IL} techniques and agents trained using TIPS. 
	Tele-operation is challenging for the demonstrator, especially for time-critical tasks such as CartPole and LunarLander where the system is inherently unstable. 
	Agents trained using IL techniques (BC and GAIL) suffer from inconsistency as well as lack of generalization of the demonstrations. 
	For the CartPole task, this problem is not as significant given the small state-action space. 
	Interactively learning via {TIPS} enables continuous improvement over time and leads to the highest performance.
	
	Figure~\ref{fig:tipsvdcoach} compares the performance of state-space ({TIPS}) and action-space ({D-COACH}) interactive learning over training episodes. The advantage of state-space feedback is significant in terms of learning efficiency for the CartPole and Reacher tasks and an increase in final performance is observed for the Reacher task.
	For the LunarLander task, no performance improvement is seen, although training with TIPS takes less time to achieve similar performance.
	While state-space feedback provides a stabilizing effect on the lander and leads to fewer crashes, participants struggle to teach it to land and thus the agent ends up flying out of the frame.
	
	\textbf{Demonstrator Task Load:} The NASA Task Load Index ratings are used to capture demonstrator task load when teaching using state-space (TIPS) and action-space (D-COACH) feedback and the results can be seen in Table \ref{table:tlxres} (Significant differences in rating are highlighted).
	
	\begin{table}[t!]
		\centering
		\captionsetup{justification=centering, font=footnotesize}
		\caption{Average ratings provided by the participants in the NASA Task Load Index questionnaire \citep{TLX}. Values are normalized, with smaller magnitude implying lower mental demand etc. (S) and (A) are used to denote state-space and action-space techniques respectively.}
		\resizebox{\textwidth}{!}
		{
			\begin{tabular}{|l  C{1.6cm} C{1.6cm} C{1.6cm} C{2.5cm} C{1.6cm} C{1.8cm}|}
				\hline
				\T & Mental&	Physical	&Temporal&	1-Performance&	Effort	&Frustration\\
				& Demand&	Demand	&Demand& &	&\\
				\hline
				\T \textit{CartPole} & & & & & &\\
				\hline
				\T TIPS (S) &	\textbf{0.29}	&0.33	&\textbf{0.33}	&0.11	&0.37	&\textbf{0.19}\\
				D-COACH (A) &\textbf{0.49}&	0.37	&\textbf{0.43}	&0.14&	0.44	&\textbf{0.3}\\
				\hline
				\T \textit{Reacher} & & & & & &\\
				\hline
				\T TIPS (S)&\textbf{0.53}	&0.64&	0.61	&0.17&	0.63&	\textbf{0.3}\\
				D-COACH (A) &\textbf{0.63}&	0.66&	0.57	&0.2&	0.61	&\textbf{0.41}\\
				\hline
				\T \textit{LunarLanderContinuous} & & & & & &\\
				\hline
				\T TIPS (S)&0.8&	0.7&	0.67	&0.3	&0.73	&0.73\\
				D-COACH (A) &0.8&	0.77&	0.67	&0.27	&0.73	&0.6\\
				\hline
			\end{tabular} 
		}
		\label{table:tlxres}
	\end{table}
	
	When teaching using {TIPS}, participants report lower ratings for the CartPole and Reacher tasks with the mental demand rating reduced by about 40\% and 16\% and participant frustration reduced by about 35\% and 25\% respectively. 
	Thus, the merits of state-space interactive learning are clear. 
	However, these advantages are task specific. 
	For the LunarLander task, demonstration in state and action-spaces is equally challenging, backed up by little change in the ratings.
	
	It is noted that actions computed based on feedback using TIPS can be irregular due to inaccuracies in model learning. 
	This was observed for the Reacher and LunarLander tasks where model learning is relatively more complex as compared to CartPole. 
	Since handling such irregular action scenarios requires demonstrator effort, this can diminish the advantage provided by state-space feedback.
	
	\subsection{Validation Tasks}
	
\begin{figure}[h!]
		\centering
		\begin{subfigure}[h]{0.4\textwidth}
			\captionsetup{justification=centering, font=footnotesize}
			\centering
			\includegraphics[scale=0.38, trim={0 0.5cm 0 1cm},clip]{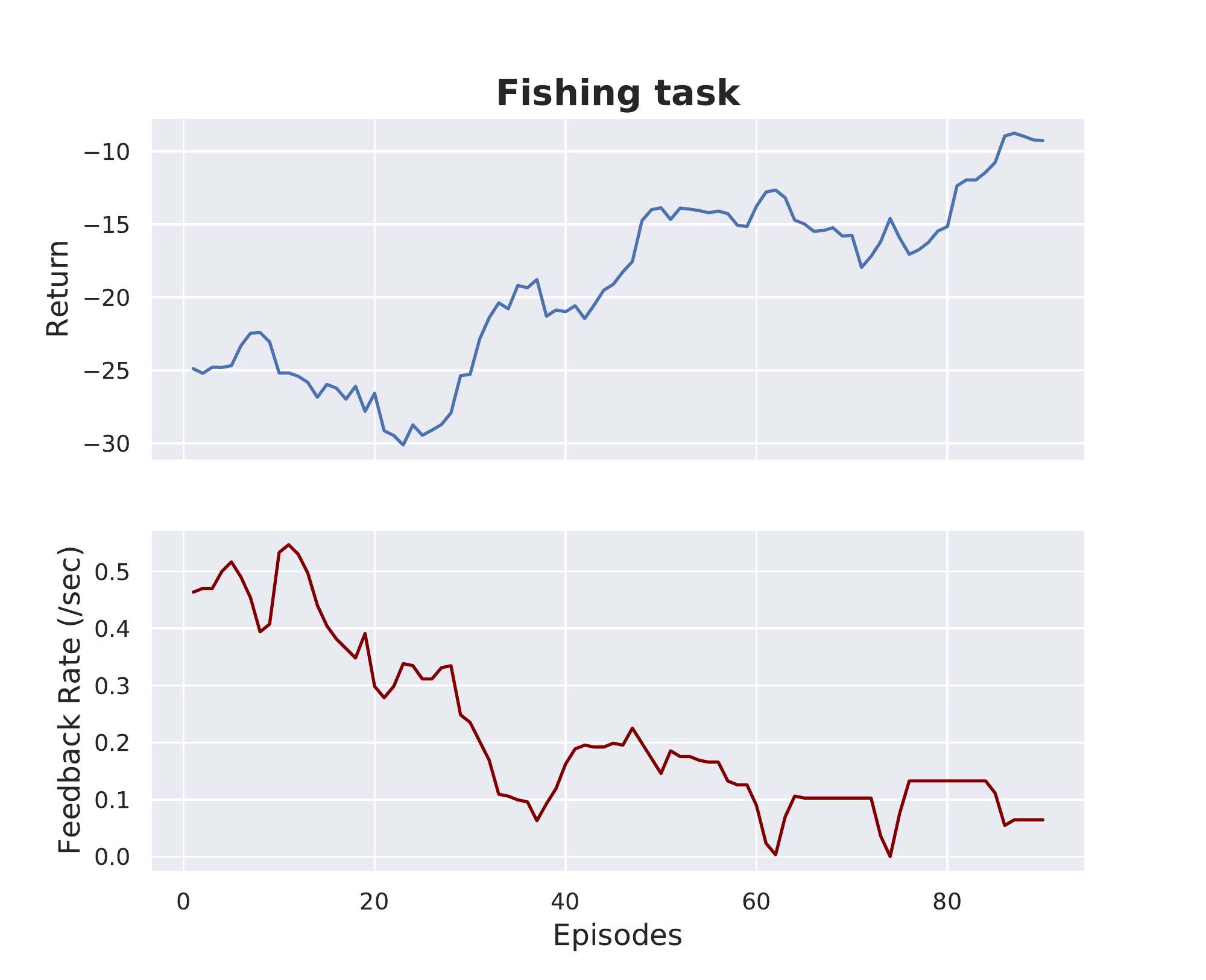}
			\label{fig:kukaplot1}
		\end{subfigure}
		\hspace*{\fill}
		\begin{subfigure}[h]{0.4\textwidth}
			\captionsetup{justification=centering, font=footnotesize}
			\centering
			\includegraphics[scale=0.38, trim={0 0.5cm 0 1cm},clip]{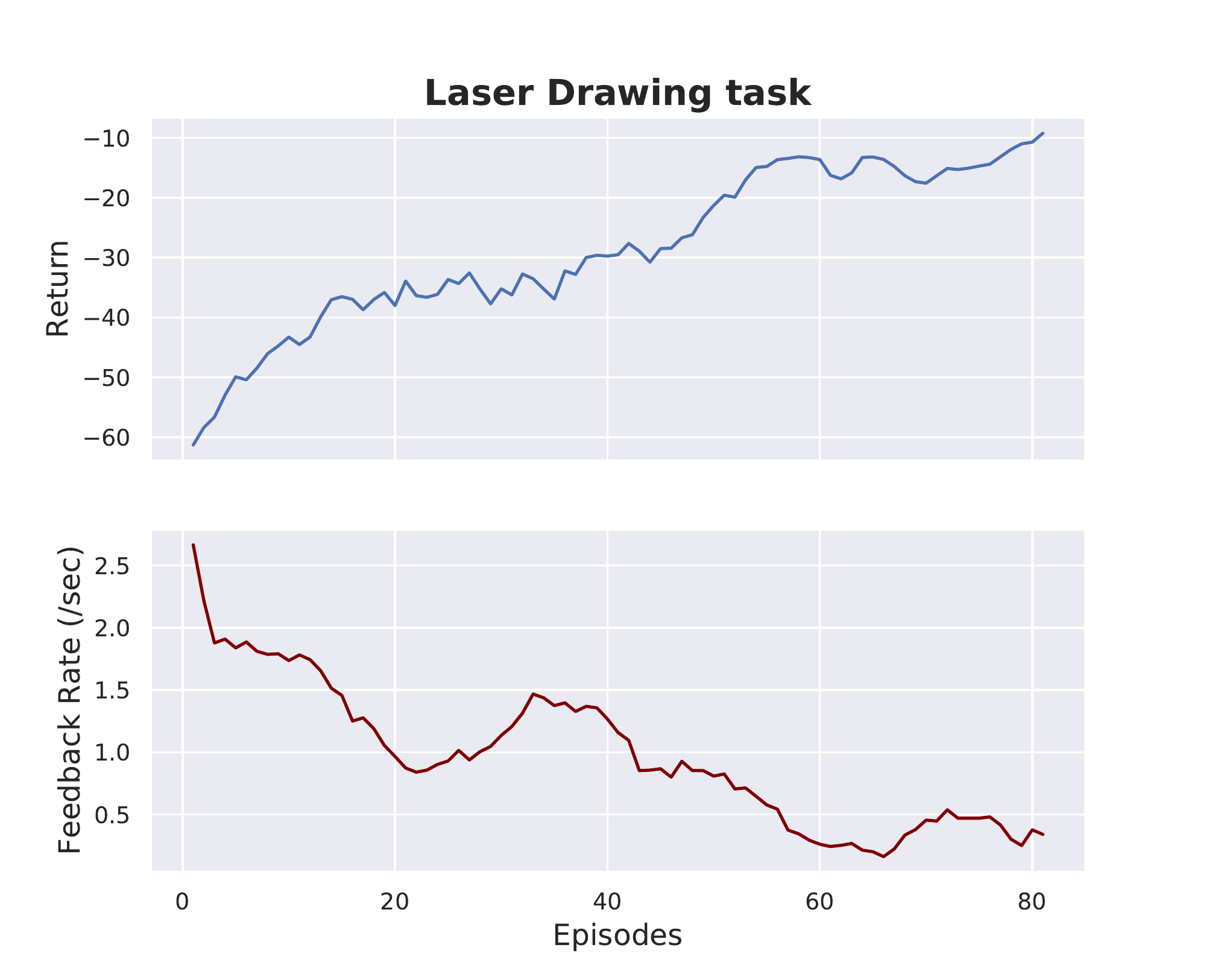}
			\label{fig:kukaplot2}
		\end{subfigure}
		\hspace*{\fill}
		\centering
		\captionsetup{justification=justified, font=footnotesize}
		\caption{Learning curves and demonstrator feedback rates for the validation experiments. Values are averaged over a rolling window of size 10. Each episode is of length 30 seconds for the Fishing task and 5 seconds for the Laser Drawing task. For the Laser Drawing task, values are averaged over learning different characters.}
		\label{fig:kukaplot}
		\vspace{-15pt}
	\end{figure}

	The agent performance and demonstrator feedback rate over learning episodes can be seen in Figure~\ref{fig:kukaplot}. 
	
	In our experiments for the Fishing task, the demonstrator's strategy is to move the end effector towards a position above the cup and choose the appropriate moment to bring the end effector down such that the swinging ball falls into the cup. 
	The agent successfully learns to reliably place the ball in the cup after 60 episodes of training (each episode is 30 seconds long).
	After about 90 episodes, the agent performance is further improved in terms of speed at which the task is completed (improvement in return from -15 to -10). 
	The feedback rate reduces over time as the agent performs better and only some fine-tuning of the behavior is needed after 60 episodes (Figure~\ref{fig:kukaplot}).
	
\begin{figure}[t!]
		\centering
		\begin{subfigure}[h]{\textwidth}
			\captionsetup{justification=centering, font=footnotesize}
			\centering
			\includegraphics[width=0.8\textwidth]{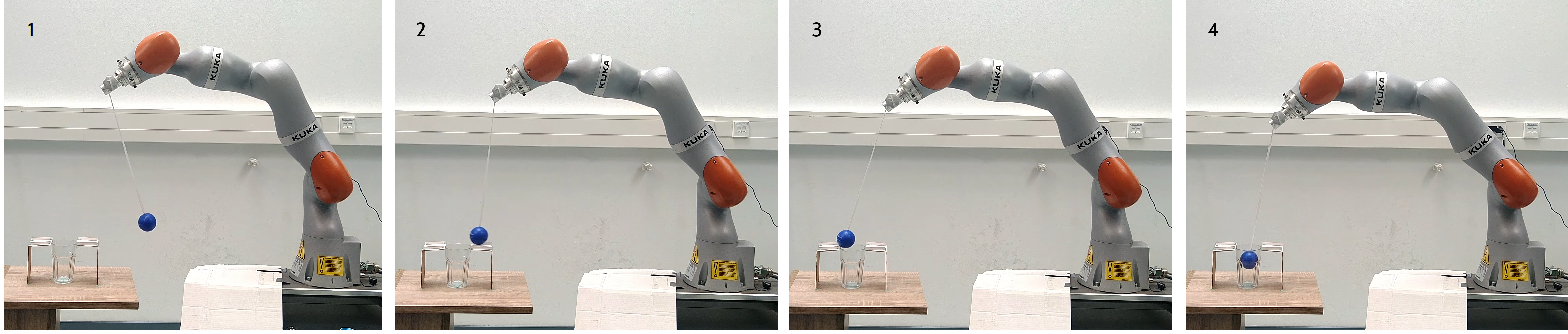}
			\caption{}
			\label{fig:kukaseq}
		\end{subfigure}
		\centering
		\begin{subfigure}[b]{0.45\textwidth}
			\vspace{5pt}
			\captionsetup{justification=centering, font=footnotesize}
			\centering
			\includegraphics[width=0.8\textwidth]{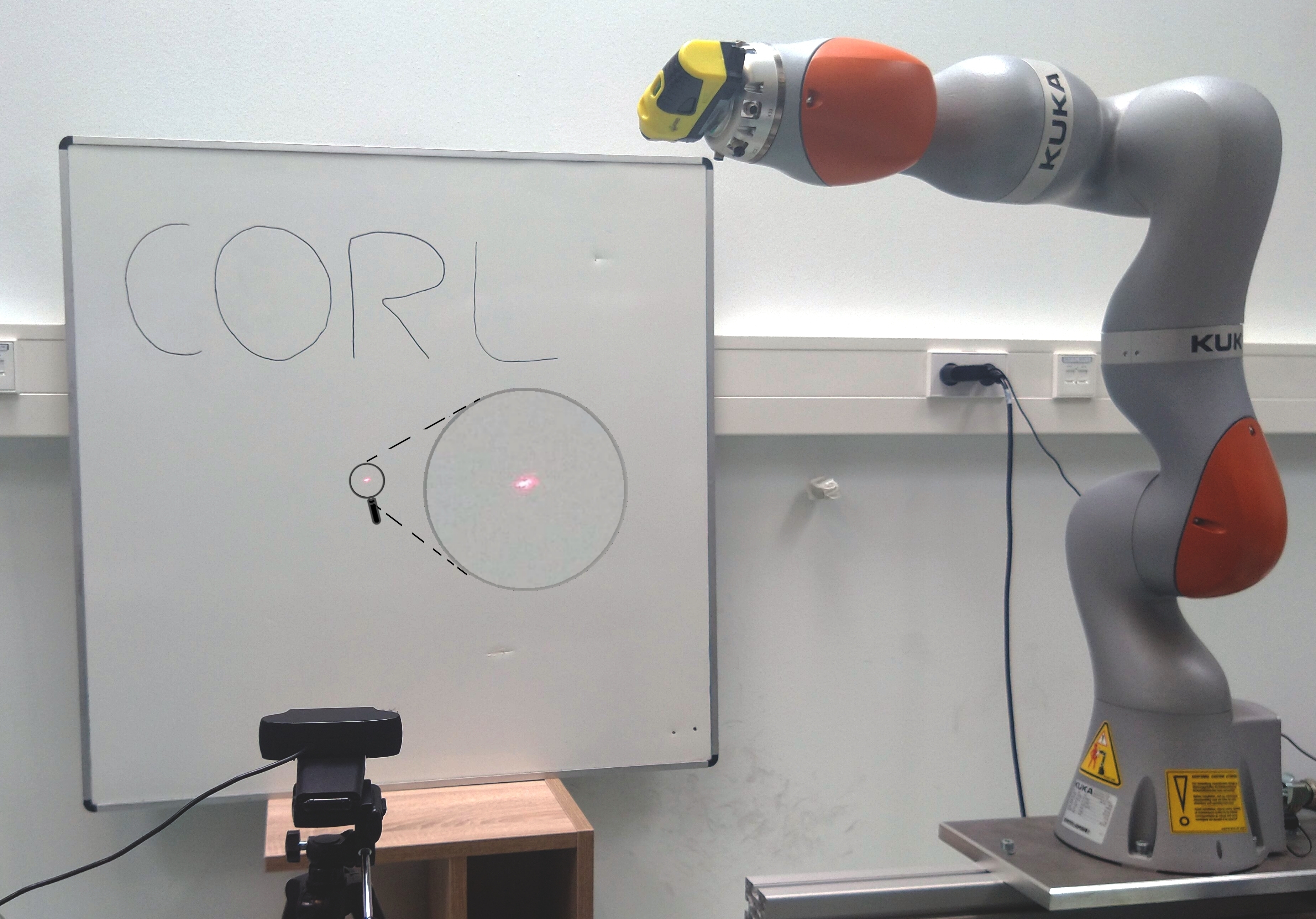}
			\caption{}
			\label{fig:kukalaser}
		\end{subfigure}
		\centering
		\begin{subfigure}[b]{0.45\textwidth}
			\captionsetup{justification=centering, font=footnotesize}
			\centering
			\includegraphics[width=0.685\textwidth]{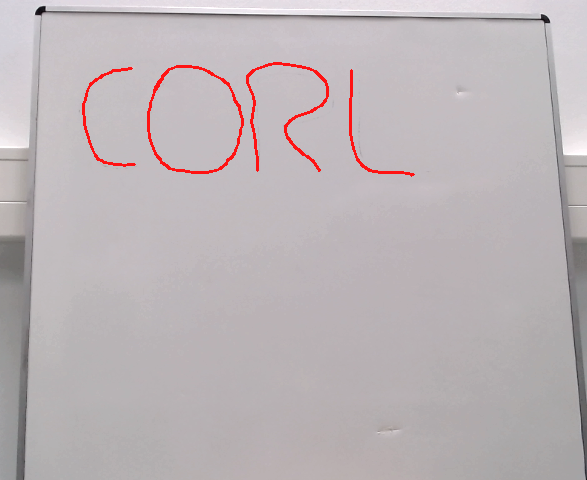}
			\caption{}
			\label{fig:kukadraw}
		\end{subfigure}
		\centering
		\captionsetup{justification=justified, font=footnotesize}
		\caption{Validation experiments with the KUKA robot. (a) Left to right, the Fishing task performed by the robot after being taught by the demonstrator for 20 minutes. (b) Representation of the Laser Drawing task. The robot is taught to move the laser point (magnified in image) to draw the characters. (c) The characters drawn by the robot (laser point trajectory tracked by the camera) after about 7 minutes of training per character.}
		\label{fig:kukaboth}
	\end{figure}

	For the Laser Drawing task, the demonstrator teaches each character separately and uses a reference drawn on the whiteboard as the ground truth. 
	The agent successfully learns to draw characters that closely resemble the reference (Figure~\ref{fig:kukadraw}) after 80 episodes of training (each episode is 5 seconds long). The feedback rate reduces over time as the basic character shape is learnt and the behavior is fine-tuned to closely match the reference character. 
	
	A video of the training and learnt behavior for both tasks is available at: \href{https://youtu.be/mKgrBgat1PM}{youtu.be/mKgrBgat1PM}.
	
	\section{Conclusion} 
	\label{sec:conclusion}
	In experiments with non-expert human demonstrators, our proposed method {TIPS} outperforms {IL} techniques such as {BC} and {GAIL} \citep{Ho2016} as well as Interactive Learning in action-space ({D-COACH}~\citep{Perez2019}). The state-space feedback mechanism also leads to a significant reduction in demonstrator task load. We have thus illustrated the viability of {TIPS} to non-expert demonstration scenarios and have also highlighted the merits of state-space Interactive Learning. Our method also has the benefit of being applicable to both continuous and discrete action problems, unlike feedback methods such as COACH \cite{Celemin2019a} (continuous actions only).
	
	To compute actions, we learn an FDM and assume no prior knowledge of dynamics. While this is advantageous in environments with dynamics that are unknown or difficult to model, learning the FDM from experience can be challenging. A lot of training data (i.e., environment interactions) may be required, else a poor model would lead to inaccurate actions being computed. A solution to this could be to use smarter exploration strategies when acquiring experience samples.
	
	Another drawback of TIPS is that the action selection mechanism requires the evaluation of samples from the entire action-space. In the relatively small dimensional spaces in our experiments, this computation was inexpensive, quick and felt instantaneous to the demonstrator. However, this does not hold for higher dimensional spaces where a lot of computational power would be required. Thus, further improvements are required to select actions in an efficient way.

	\acknowledgments{
		This research has been funded partially by the ERC Stg TERI, project reference \#804907. We would like to thank Rodrigo P{\'e}rez-Dattari for his comments and suggestions. We would also like to thank the CoRL reviewing committee for their insights which helped improve the final content of the paper.} 
	
	
	
	\bibliography{paper}  
	\clearpage
	\section*{\centering Supplementary Information On Experiments}
	
	We implemented TIPS in Python and used the TensorFlow Python library~\citep{tensorflow2015} to train the neural networks for the forward dynamics model and agent policy. Our implementation is available at \href{https://github.com/sjauhri/Interactive-Learning-in-State-Space}{github.com/sjauhri/Interactive-Learning-in-State-Space}.
	
	We ran experiments 
	to evaluate our method TIPS in simulated OpenAI Gym~\citep{Brockman2016} environments and to validate it on two manipulation tasks with a KUKA LBR iiwa robotic arm. In all the experiments, the demonstrator's input was taken via arrow keys on a keyboard. For the validation experiments with the robotic arm, we used the iiwa stack \citep{Hennersperger2017} to interface with the robot using ROS commands. Thus, actions in the policy were in the form of joint velocity commands sent to the robot. The frequency of actions, i.e., the controller frequency was set to 10 Hz. The state-space for the tasks included the robot joint positions, velocities along with the camera-tracked position and velocity of the ball (in the Fishing task) or the position of the laser point (in the Laser Drawing task). The experiments were first tested in simulations in Gazebo followed by execution using the real robot.
	A video of the training and learnt behavior for both validation tasks is available at: \href{https://youtu.be/mKgrBgat1PM}{youtu.be/mKgrBgat1PM}.
	
\end{document}